\documentclass[letterpaper,10pt,conference]{ieeeconf}
\IEEEoverridecommandlockouts
\makeatletter
\let\NAT@parse\undefined
\makeatother

\usepackage{amsmath,amssymb,amsfonts}
\usepackage{booktabs}
\usepackage[font=footnotesize,labelformat=simple]{subcaption}
\usepackage[font=footnotesize]{caption}

\usepackage{tikz}
\usepackage[mode=buildnew]{standalone}
\usepackage{xspace}
\usepackage{cite}
\usepackage[noend]{algorithmic}
\usepackage{algorithm}
\usepackage{caption}
\usepackage{subcaption}
\usepackage{tabulary}
\usepackage{verbatim}
\usepackage[numbers,sort&compress]{natbib}
\usepackage{thmtools}
\usepackage{mathtools}
\usepackage{multirow}
\usepackage{flushend}
\usepackage{hyperref}
\usepackage{enumitem}
\usepackage{dsfont}
\usepackage[letterpaper, left=0.75in, right=0.75in, bottom=0.75in, top=0.75in]{geometry}

\hypersetup{
  pdfinfo={
    Title={BDMP},
    Author={},
    Subject={Motion planning},
    Keywords={Motion and Path Planning; Mobile Robots}
  },
  bookmarksopen=true,
  bookmarksnumbered=true,
  pdfpagemode=UseOutlines,
  colorlinks=true,
  linkcolor=blue,
  anchorcolor=blue,
  citecolor=blue,
  filecolor=blue,
  menucolor=blue,
  urlcolor=blue
}

\title{\LARGE \bf Multi-Sample Long Range Path Planning under Sensing Uncertainty for Off-Road Autonomous Driving}

\author{
  Matt Schmittle, Rohan Baijal, Brian Hou, Siddhartha Srinivasa, Byron Boots%
  \thanks{
    Distribution Statement A (Approved for Public Release, Distribution Unlimited).
  
    All authors are with the
    Paul G. Allen School of Computer Science \& Engineering,
    University of Washington
    \texttt{\{schmttle, rbaijal, bhou, siddh, bboots\}@cs.washington.edu}}%
}

\usepackage{amsmath,amsfonts,bm, bbm}

\def\eqref#1{equation~\ref{#1}}
\def\Eqref#1{Equation~\ref{#1}}

\def\1{\bm{1}}

\DeclareMathAlphabet{\mathsfit}{\encodingdefault}{\sfdefault}{m}{sl}
\SetMathAlphabet{\mathsfit}{bold}{\encodingdefault}{\sfdefault}{bx}{n}

\newcommand{\real}[0]{\mathbb{R}}

\newcommand{\xxnote}[3]{}
\ifx\hidenotes\undefined
  \renewcommand{\xxnote}[3]{\color{#2}{#1: #3}}
  \definecolor{SeaGreen}{HTML}{3FBC9D}
  \definecolor{Orange}{HTML}{D95F02}
  \definecolor{RacingGreen}{HTML}{004225}
\fi

\newcommand{\sref}[1]{Section~\ref{#1}} 

\begin{document}
\maketitle
\thispagestyle{empty}
\pagestyle{empty}

\begin{abstract}

We focus on the problem of long-range dynamic replanning for off-road autonomous vehicles,
where a robot plans paths through a previously unobserved environment while continuously receiving noisy local observations.
An effective approach for planning under sensing uncertainty is determinization,
where one converts a stochastic world into a deterministic one and plans under this simplification.
This makes the planning problem tractable, 
but the cost of following the planned path in the real world may be different than in the determinized world.
This causes collisions if the determinized world optimistically ignores obstacles,
or causes unnecessarily long routes if the determinized world pessimistically imagines more obstacles.

We aim to be robust to uncertainty over potential worlds while still achieving the efficiency benefits of determinization.
We evaluate algorithms for dynamic replanning on a large real-world dataset of challenging long-range planning problems from the DARPA RACER program.
Our method, Dynamic Replanning via Evaluating and Aggregating Multiple Samples (DREAMS), outperforms other determinization-based approaches
in terms of combined traversal time and collision cost. \href{https://sites.google.com/cs.washington.edu/dreams/}{https://sites.google.com/cs.washington.edu/dreams/}
\end{abstract}

\section{Introduction}
\label{sec:introduction}

Inspired by the DARPA RACER program~\cite{racer}, we focus on the problem of motion planning under sensing uncertainty for autonomous off-road vehicles travelling over tens of kilometers. RACER challenges teams to program an autonomous off-road vehicle equipped only with onboard sensing and compute to navigate complex terrain (deserts, forests, hills) over long distances. Unlike on-road driving (which contends with lanes, signs, and rules of the road), off-road driving has much less structure. The robot can go wherever it can effectively traverse, posing a unique robotics challenge.

This flexibility relies on the autonomy system's onboard sensors and perception system to discern what terrain is and is not traversable. For example, terrain that is far from the robot may be difficult to classify precisely due to e.g.,natural occlusions (hills, trees), few sensor readings, or lack of training examples in the given environment.
This noisy perception creates both false obstacles and false freespace;
a downstream planning algorithm that is unaware of this uncertainty may produce dangerous collision-bound or roundabout paths.

\begin{figure}[!t]
\centering
\includegraphics[width=0.6\linewidth]{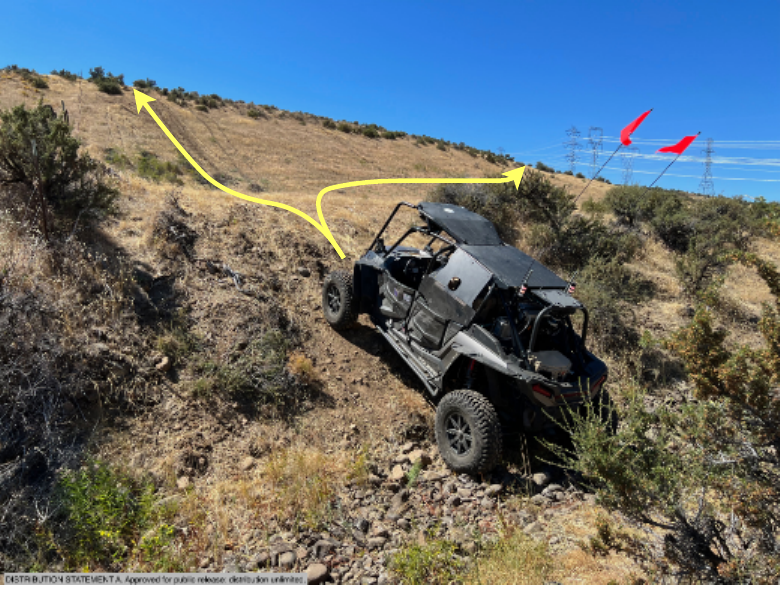}
\caption{An autonomous off-road vehicle's long-range planner needs to decide the best way up a hill, given blind spots and imperfect sensing.}
\label{fig:cover}
\vspace{-1\baselineskip}
\end{figure}

Thus, uncertainty is the core challenge of long-range planning: an algorithm must appropriately consider this sensing uncertainty from perception when planning paths (Fig. \ref{fig:cover}).

In this setting, a robot plans paths through a previously unobserved and potentially hazardous environment,
while continuously receiving and reacting to noisy local observations.

This can be cast as a Partially Observable Markov Decision Process (POMDP).
Solving a POMDP is PSPACE-Complete~\cite{papadimitriou1987}, so we frame this as a Bayesian Dynamic Motion Planning Problem (BDMP) to impose structure and make the problem tractable.
The BDMP problem differs from a POMDP in that uncertainty originates only from the robot’s ignorance about the environment and the agent maintains a posterior over possible environments given its observations. Given the environment, the transition, the reward function and robot's internal state are fully observable~\cite{hou2023}.

\begin{figure*}[!t]
\centering
\includegraphics[width=0.9\linewidth]{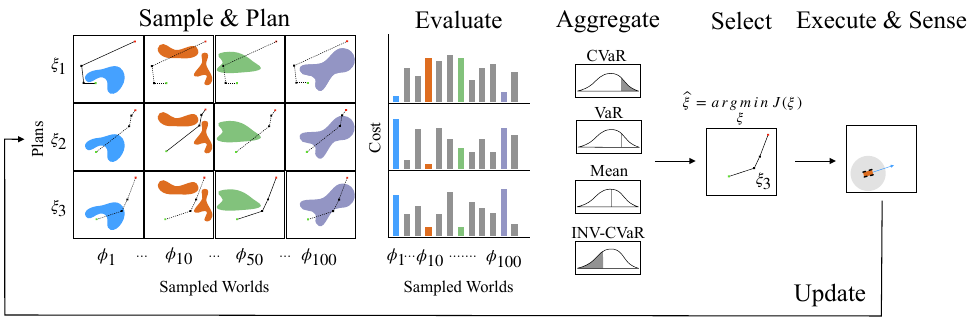}
\caption{
Overview of DREAMS.
\emph{Sample \& Plan}: Sample many worlds from the posterior distribution, and plan the optimal path on a subsample of worlds ($\phi_1$, $\phi_{10}$, $\phi_{50}$ above).
\emph{Evaluate}: Evaluate the cost of each resulting plan against the full set of sampled worlds.
\emph{Aggregate}: Aggregate the resulting cost distribution with a summary statistic (e.g., mean or CVaR)
\emph{Select}: Select the plan with minimal aggregated cost.
}
\label{fig:summary}
\vspace{-1\baselineskip}
\end{figure*}
Previous works have exploited this structure via the framework of determinization in the face of uncertainty---repeatedly solving and executing relatively-inexpensive determinized planning problems---with strong theoretical and practical results~\cite{hou2023,somani2013,lim2017,yoon2007}.
Although determinization is computationally efficient, it does not consider what happens when the determinized problem that it solved diverges from reality.
This can manifest as optimism (causing collisions) or pessimism (causing roundabout paths, or no path at all). 

Our key insight is that this deficiency stems from determinization's limited ability to \emph{reason about the distribution of costs over plausible worlds}.
Thus, we leverage multi-sample posterior sampling to reap some of the computational benefits of determinization
while preserving the planner's ability to reason across multiple plausible environments.
Our resulting multi-sample determinization algorithm, Dynamic Replanning via Evaluating and Aggregating Multiple Samples (DREAMS),
considers multiple plausible optimal paths and multiple plausible worlds.
With this framework, DREAMS enables reasoning not just over the distribution of worlds, but also over additional parameters such as traversal speed.
We show that with the correct instantiation, DREAMS outperforms prior determinization strategies on realistic long-range off-road robot navigation tasks.

We make the following contributions:
\begin{itemize}
\item We introduce DREAMS (Fig. \ref{fig:summary}), an algorithm for planning under uncertainty
    that maintains the ease of planning from determinization
    while also being able to consider uncertainty over worlds in decision making.
\item On a large dataset of challenging long-range planning problems, we demonstrate that DREAMS plans effectively under uncertainty to achieve lower total cost compared to other determinization methods.
\end{itemize}
\section{Related Work}
\label{sec:related-work}

There has been a fair amount of prior work on dynamic replanning under uncertainty. D* and D* Lite are well-known dynamic replanning search algorithms that have been demonstrated to quickly replan in real-world settings~\cite{stentz1994,koenig2002,ferguson2005}. Neither are designed to reason about uncertainty and are optimistic about unknown parts of the environment. Re-planning is triggered when the planned path is deemed in collision. In an environment with no sensing uncertainty, this is effective because collisions can be easily determined. In an uncertain environment, D* must either collide with an obstacle to detect a collision or blindly trust its noisy sensors. That being said, D*'s efficient re-use of the search tree could be incorporated into methods that consider environment uncertainty directly.

The BDMP problem can be framed as a POMDP with unknown state, but known transitions and rewards. POMDPs have a plethora of approaches but a notable few rely on a fixed or sampled set of MDPs to make the problem more tractable~\cite{ross2008, silver2010, somani2013, sunberg2017, chen2016}. Most similar to our approach is DESPOT~\cite{somani2013}, which samples a set of K scenarios and builds a tree to alleviate the curse of dimensionality. While promising, solving a POMDP is PSPACE-complete~\cite{papadimitriou1987} and this work instead focuses on a tractable MDP relaxation with discrete set of states and known transitions represented as a graph (instead of a tree) with unknown reward.

The Canadian Traveler's Problem~\cite{papadimitriou1989} and specifically its stochastic variants~\cite{eyerich2010, dey2014, lim2017, yoon2008} are most similar to the BDMP setting. In both cases, edge collisions are discovered when the agent reaches an incident vertex. This setting follows our observations from real mobile robotic systems: sensing is more accurate near the robot. The stochastic variants additionally use this information to update unobserved edge probabilities. Our setting is similar except we discover the true cost of an edge when we traverse it, and we additionally incorporate limited range noisy sensing.

Risk-aware planning can be seen as another form of planning under uncertainty~\cite{chung2019, barbosa2021, suresh2019, feyzabadi2014, murphy2013, cai2022}. Barbosa et al.~\cite{barbosa2021} converts the popular Conditional Value at Risk (CVaR)~\cite{uryasev2001} metric into a cost function for planning and accepts new plans only when risk increases along the current trajectory. While these methods have shown great promise, none have investigated planning under the setting of onboard sensing where uncertainty increases further from the robot. Therefore, in this work we compare with these approaches as a baseline (\sref{direct}).  

Determinization, making a deterministic approximation of a stochastic problem, has been effective for planning under uncertainty~\cite{yoon2007}. In particular, a variety of works~\cite{hou2023, asmuth2012, strens2000, wilson2007, chung2019} have used posterior sampling~\cite{thompson1933} to make the planning problem tractable and only require sample access to the posterior. Dynamic Replanning with Posterior Sampling (DRPS)~\cite{hou2023} samples one problem and solves it optimally, letting sampling naturally balance exploration and exploitation. A key observation from DRPS is that gaining information from the world as a mobile robot is relatively easy without explicit exploration. As the robot moves, it easily gains more sensor information to clarify its future actions. Our work leverages this insight and we apply posterior sampling as a determinization strategy. Sampled A*~\cite{chung2019} further utilizes multiple samples per replan and accepts the most likely path, showing promising results. DREAMS also uses multiple samples, but differs in how it selects the plan to follow by considering a distrbution of costs. We compare DRPS and Sampled A* with DREAMS in a realistic setting where sensing is noisy and there is a limited observation range.
\section{Bayesian Dynamic Motion Planning with Collisions}
\label{sec:problem}
Given a start state $x_s$ and goal state $x_g$ in configuration space $\mathcal{X}$ and a set of environments $\Phi$, we seek to minimize the expected total time of traversing from start to goal under the distribution of environments $P(\phi)$. To help solve this problem, we are given a measure of uncertainty over environments modeled as the posterior distribution $P(\phi | \psi_t)$ where $\psi_t$ is the history of observations from time $[0,t]$. This problem extends the Bayesian Dynamic Motion Planning Problem (BDMP)~\cite{hou2023} to consider the added cost of potential collisions. Off-road collisions can be dangerous---injuring the rider or damaging the robot---and require additional time to recover from.

Similar to DRPS, we focus on planning over roadmaps. Specifically, we are given a graph $G$ with vertices $V$ and edges $E$. Each edge has a traversal time $w: E \rightarrow \real^+$ and a collision status $\phi(e)$ where $\phi(e) = 1$ means $e$ is a collision-free edge in world $\phi$. A path $\xi_t= (e_1, e_2,\dots,e_t)$ is defined as a sequence of edges. Since we are in a dynamic setting, traversing edges adds observations to our history $\psi_t$.

In the motivating RACER scenario, the goal is to get from start to goal as fast as possible. Thus, we consider the following two metrics: Traversal Time and Collision Cost. We additionally consider collision cost because optimizing for traversal time alone can lead to impractical algorithms that collide frequently with obstacles.
\begin{itemize}
    \item \textbf{Traversal Time.} $T(\xi) = \sum_{e\in \xi} w(e)$ is the time it takes the robot to traverse to the goal. Edges in collision are still counted toward traversal time.
    \item \textbf{Collision Cost.} $C(\xi; \rho) = \sum_{e\in \xi}\mathds{1} (\phi(e) = 0) c(e) \rho(e)$.
    $c(e)$ is the collision cost and $\rho(e)$ adjusts the relative cost of a collision compared to traversal time.
\end{itemize}
The total cost to reach the goal in one planning episode is:
\begin{equation}\label{eq:1}
    J(\xi; \rho) = T(\xi) + C(\xi; \rho)
\end{equation}
where $\rho(e) = \alpha$ is a constant.

\section{Proposer-Acceptor Approach}
\label{sec:approach}

To summarize various approaches from the literature,
we decompose algorithms into a \emph{proposer} and an \emph{acceptor}. The proposer proposes a set of paths $\Xi =\{ \xi^{0}, \xi^{1}, \dots, \xi^{n} \}$. The acceptor considers the proposed paths and accepts one. The robot then follows the accepted path for a step, receives observations, and updates the posterior distribution. It then replans and repeats until the goal is reached. 

\begin{figure}[t!]
    \centering
    \includegraphics[width = 0.9\linewidth]{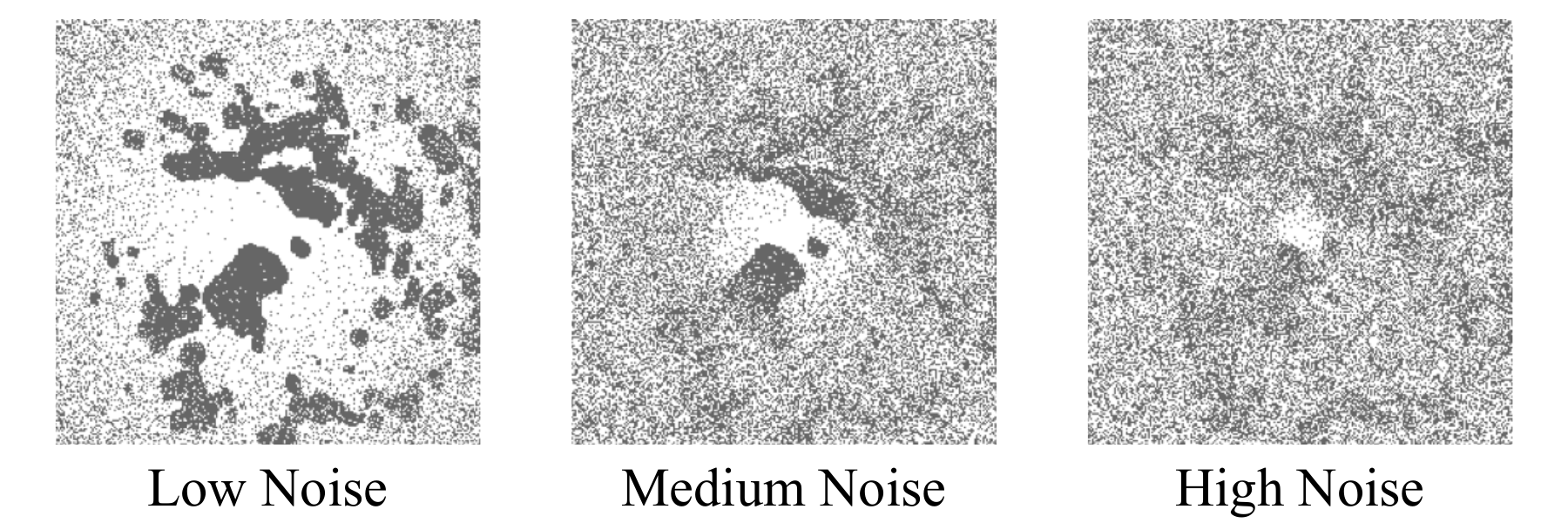}
    \caption{
The sensor noise levels used in testing: $\eta_{\mathrm{low}} = 10^{-4}$, $\eta_{\mathrm{med}} = 10^{-3}$, $\eta_{\mathrm{high}} = 10^{-2}$.
Our simplified noise model defines the probability of receiving a correct observation for a query point distance $d$ away as $\max ( \exp(-\eta d^2), p_{\mathrm{min}})$.
The minimum probability threshold $p_{\mathrm{min}}$ is set to 0.6 to provide some signal at the edge of the robot's observation range;
note that the minimum possible value of $p_{\mathrm{min}}$ for binary occupancy is 0.5 (pure noise).
With these parameters, the robot receives approximately 96\%, 72\%, and 61\% correctly observed pixels per observation.
}
    \label{fig:noise}
    \vspace{-1\baselineskip}
\end{figure}

\subsection{DREAMS Proposer}
The DREAMS proposer is based on posterior sampling, where at each step we are sampling from the distribution of optimal plans $P(\xi^* | \psi_t) = P(\xi | \phi) P(\phi | \psi_t)$. This is achieved via sampling from the posterior over worlds $P(\phi | \psi_t)$ and then planning the optimal path on each sampled world (Fig. \ref{fig:summary}, \emph{Sample \& Plan}). As the robot learns more about the world, this process naturally exploits the knowledge we gain by reducing the spread of the distribution of worlds/plans. Similar to Sampled A*, DREAMS samples multiple plans to approximate the distribution $P(\xi^* | \psi_t)$.

\begin{figure*}[!t]
    \centering
    \includegraphics[width=\linewidth]{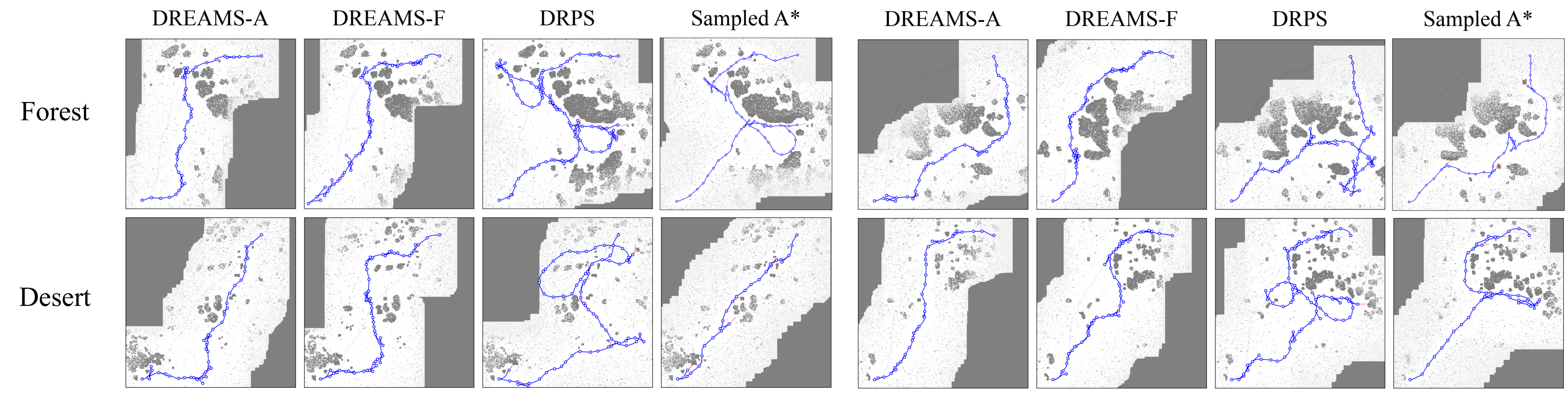}
    \caption{Traversed paths of each algorithm (blue edges) on two example worlds from each of the Forest and Desert datasets, with the same world in each set of four. Each algorithm receives observations with high noise, and is penalized with a collision factor of $\alpha=10$. With high noise, DRPS frequently backtracks and changes direction while Sampled A* incurs many collisions (red edges). Both DREAMS variants follow more direct paths without collisions.}
    \label{fig:qualitative}
    \vspace{-1\baselineskip}
\end{figure*}

\subsection{DREAMS Acceptor}
Unlike prior determinization approaches, we evaluate the cost of each sampled plan against a distribution of sampled worlds (Fig. \ref{fig:summary}, \emph{Evaluate}). The sampled worlds do not need to be the same as the ones used for planning. Empirically, we have found that planning is typically the bottleneck rather than evaluation. Therefore, we opt to sample many more worlds for evaluation than planning.

For each plan, we compute a summary statistic for this resulting distribution of costs (Fig. \ref{fig:summary}, \emph{Aggregate}) and select the plan that minimizes aggregate cost. Depending on the application, the summary statistic can vary. For example, selecting the minimum cost for a given plan is an optimistic strategy that looks at a plan's cost under the best-case scenario. CVaR summary statistics balance the risk of high cost paths (in this case, collision-prone paths) more carefully.

This approach is extremely flexible. 
In Section \ref{sec:experiments}, we additionally use this acceptor to explore different velocity profiles to further reduce expected cost.
We now describe example evaluation and aggregation functions, although these design choices will vary based on the application.

\subsubsection{DREAMS Evaluation Function}
We make a small modification to Eq.~\ref{eq:1} to use as the DREAMS evaluation function.
\begin{align}\label{eq:2}
    \widehat{J}(\xi) &= T(\xi) + C(\xi; \tau) \\
    \tau(e) &= \mathds{1}(e = e_0)\alpha + \mathds{1}(e \neq e_0)
\vspace{-1\baselineskip}
\end{align}
Because DREAMS plans in a receding-horizon fashion,
$\widehat{J}(\xi)$ considers the collision factor $\alpha$ only on the immediate edge and assigns future potential collisions a relative cost of 1.
Equally weighting all collisions can create overly conservative behavior due to noisy observations farther from the robot.
Reducing future collision cost helps avoid these scenarios.

\subsubsection{DREAMS Aggregation Function}
We optimistically take the mean of the best 75\% of the distribution of costs, calling it the Inverse CVaR. This reduces the effect of unlikely high-cost outliers.
Like CVaR it also considers the width of the distribution:
as increased uncertainty causes the cost distribution to spread out, the increased aggregate cost promotes caution.

\subsection{Benchmark Overview}
We briefly summarize each benchmark algorithm with this framework.
Each algorithm proposes plan(s), accepts a plan, follows one edge, and replans.
\begin{itemize}
    \item \textbf{DRPS~\cite{hou2023}.}
    Proposer: Sample one plan from the posterior. Acceptor: choose only plan.
    \item \textbf{Sampled A*~\cite{freeman1977}.}
    Proposer: Sample multiple plans from the posterior.
    Acceptor: choose the most likely plan.
    Determined by computing edge centrality across plans and accepting the plan with maximum mean edge centrality.
    \item \textbf{Direct.}\label{direct}
    Proposer: Compute the plan that minimizes risk-aware evaluation cost in expectation.
    Acceptor: choose only plan.
    \begin{equation}\label{eq:5}
        \mathop{\mathds{E}}[\widehat{J}(\xi)]  = \sum_{e \in \xi} w(e) + P(\phi(e) = 0)c(e) \tau(e)
    \end{equation}
\end{itemize}

\section{Experiments}
\label{sec:experiments}

The following simulation experiments are designed to replicate an off-road autonomous driving scenario where the planner faces limited sensor range and noisy perception. The robot must navigate as efficiently as possible while avoiding dangerous collisions, re-planning at at each step.

We compare DREAMS to DRPS and Sampled A*, as both incorporate posterior sampling and present strong results on similar problems.
DRPS highlights the difference between single and multi-sample posterior sampling,
while Sampled A* compares the evaluation/aggregation acceptor strategy with an approximate MAP estimate.
These baseline algorithms only consider the geometric path at an arbitrary fixed speed.
We include results for DREAMS-Fixed to compare most directly with these algorithms,
and provide additional results for DREAMS-Adaptive to demonstrate our ability to evaluate paths under different parameters in this case speed.
Finally, we evaluate a benchmark that optimizes \Eqref{eq:2} directly (Direct) to demonstrate the benefits of posterior sampling.

We consider the following hypotheses:
\begin{enumerate}[label=\textbf{H\arabic*.}]
\item
    \textit{Both DREAMS variants will incur lower total cost compared to DRPS and Sampled A*.}
    More sampled plans will help DREAMS reduce cost variance relative to DRPS.
    Reasoning about a distribution of costs rather than accepting approximate MAP estimate will help DREAMS incur less collision cost than Sampled A*.
\item
    \textit{DREAMS-Adaptive will reduce collision cost and total cost compared to DREAMS-Fixed.}
    Reasoning about distribution of speeds allows the vehicle to slow down when the likelihood of a collision increases and speed up when the path is likely free.
\item
    \textit{Increasing the number of sampled plans will reduce the total cost (with diminishing returns).}
    More plans will provide more options to evaluate, until all likely options are enumerated.
\item
    \textit{Increasing the number of sampled worlds considered during evaluation will reduce the total cost (with diminishing returns).}
    More world samples from the distribution will better estimate the distribution.
\item
    \textit{DREAMS will incur lower total cost compared to Direct.}
    To plan directly with the cost function, Direct replaces the indicator in \Eqref{eq:2} with a probability.
    This can result in extremely unlikely plans under the posterior $P(\xi^*|\psi_t)$,
    which DREAMS probabilistically avoids by construction.
\end{enumerate}

\subsection{Experimental Setup}

\begin{figure*}[ht!]
    \centering
    \includegraphics[width=0.9\textwidth]{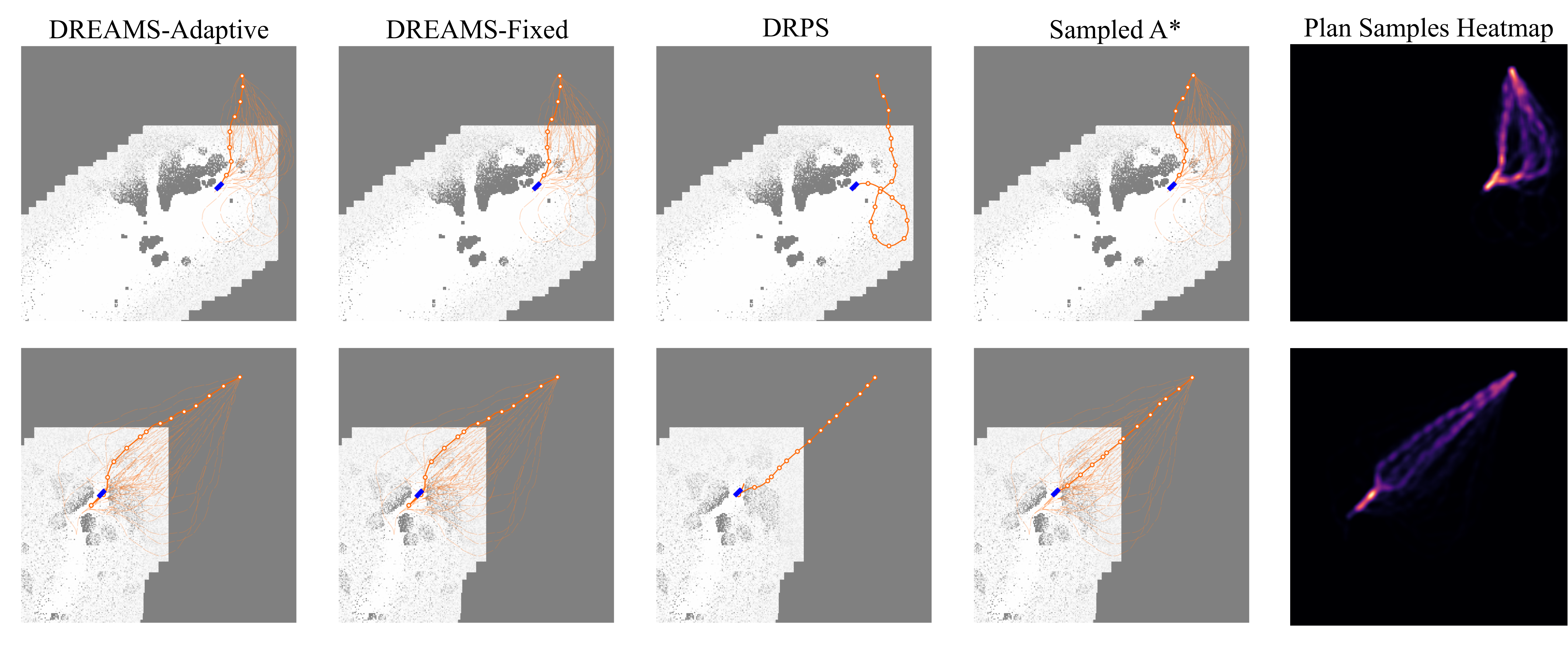}
    \caption{Qualitative comparison of each approach, given the exact same sampled worlds and paths.
    Robot (blue), proposed paths (light orange), accepted path (bright orange).
    \emph{Top:} All plans except DRPS find a path through the gap. DRPS happened to sample a world that did not fit through the gap, producing a longer route. 
    \emph{Bottom:} All plans except Sampled A* reverse from a likely obstacle in front of the robot. Sampled A* does not explicitly consider the cost of collision and accepts a path going through the obstacle. 
    \emph{Right:} Looking at the heatmap, areas where more plans overlap are hotter. 
    As there is little overlap besides the start position, Sampled A* has less signal to choose the most likely plan; its decision is almost a uniform random sample.}
    \label{fig:comparison}
    \vspace{-1\baselineskip}
\end{figure*}

\subsubsection{Real World Occupancy Grids and Speeds}

We evaluate performance with two real-world datasets of long-range planning problems through open desert environments ($N=83$) and more challenging crowded forest environments ($N=51$). These datasets were collected through the RACER program.

 Worlds are $100 \times 100$ meters at a resolution of 0.4 m/px. The robot moves on a graph covering the space at speeds ranging from 1--10 m/s. As the robot traverses, it receives observations at 1 Hz.
Therefore, speed affects both traversal time and the number of observations received. The robot can move in reverse at a fixed speed of 1 m/s. For all approaches, we discourage reversing by prompting each planning call to find a solution without reversing. If unsuccessful, it retries with reversing allowed. 
We evaluate on ten random seeds for each world $\phi$, noise level $\eta$, and collision $\alpha$.

\subsubsection{Limited Range Noisy Observations}
It is a popular choice among autonomous vehicles to process raw sensor input into semantic classification of the environment using a deep neural network~\cite{shaban2022, schmid2022}. Noisy sensors and limited training introduce uncertainty into the resulting semantic segmentations. Generally, the predictions become noisier farther away from the robot where there is less (and noisier) sensor information. Predicted semantics eventually become too noisy and are thus limited to a reliable range. To simulate this, we add a limited range observation module that simulates classification of obstacles. The robot can only observe a patch of $50 \times 50$ meters centered around itself. Fig. \ref{fig:noise} describes the sensor model within this limited range observation.

Outside of the limited range observation, we optimistically assume that the space is free to allow posterior sampling to discover many plausible paths. This is similar to how real-world systems work where unknown space is a fixed cost. The posterior is updated using Bayes' Rule. 





\subsubsection{Posterior Sampling}
To sample from the posterior distribution over optimal plans $P(\xi^* | \psi_t)$,
we sample from the posterior distribution over worlds $P(\phi | \psi_t)$ and plan on each world.
We sample a world by sampling over the posterior distribution of roadmap edges,
created by taking the maximum posterior collision probability across all pixels marked by the robot's swept volume ($3.5 \times 1.5$ meters) along an edge.

\subsubsection{Planning and Execution Parameters}
For DREAMS and Sampled A*, we choose to plan with 100 posterior samples using A*. DREAMS evaluates each plan according to (\Eqref{eq:2}) against $10^4$ sampled worlds. DREAMS-Adaptive additionally considers multiple speed profiles. For each sampled plan, it creates five timed trajectories each with a separate speed profile. Profiles are all 5 m/s within the observed area and optimistically 10 m/s outside the observed area, but differ in the first edge traversal speed $\{1, 3, 5, 7, 10\}$ m/s. DREAMS-Adaptive executes the chosen speed for the chosen plan for one time step before re-planning. DREAMS-Fixed and the other algorithms consider only one speed profile: 5m/s in observed area, 10m/s outside observed area.

The collision cost is proportional to the robot's pre-collision speed (\Eqref{eq:1}), as higher speed collisions are more dangerous.
We vary $\alpha$ across $\{1, 10, 20\}$ to characterize performance with different relative collision costs.

\subsubsection{Metrics}
We compare each algorithm's incurred cost to the cost incurred by an oracle $\xi^{opt}$,
which has full information about the world and traverses at 10 m/s without collisions.
\begin{equation}\label{eq:subopt}
  \mathrm{Suboptimality}
  = J(\xi)/ J(\xi^{opt})
  = J(\xi)/ T(\xi^{opt})
\end{equation}

\begin{figure*}[ht!]
     \centering
     \begin{subfigure}[t]{\linewidth}
        \centering
        \includegraphics[width=\linewidth]{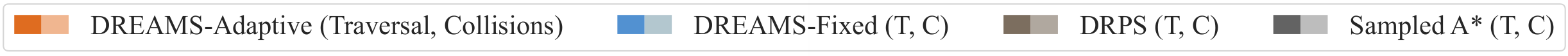}
     \end{subfigure}
     \begin{subfigure}[t]{0.49\linewidth}
        \begin{subfigure}[t]{\linewidth}
            \centering
            \includegraphics[width=\linewidth]{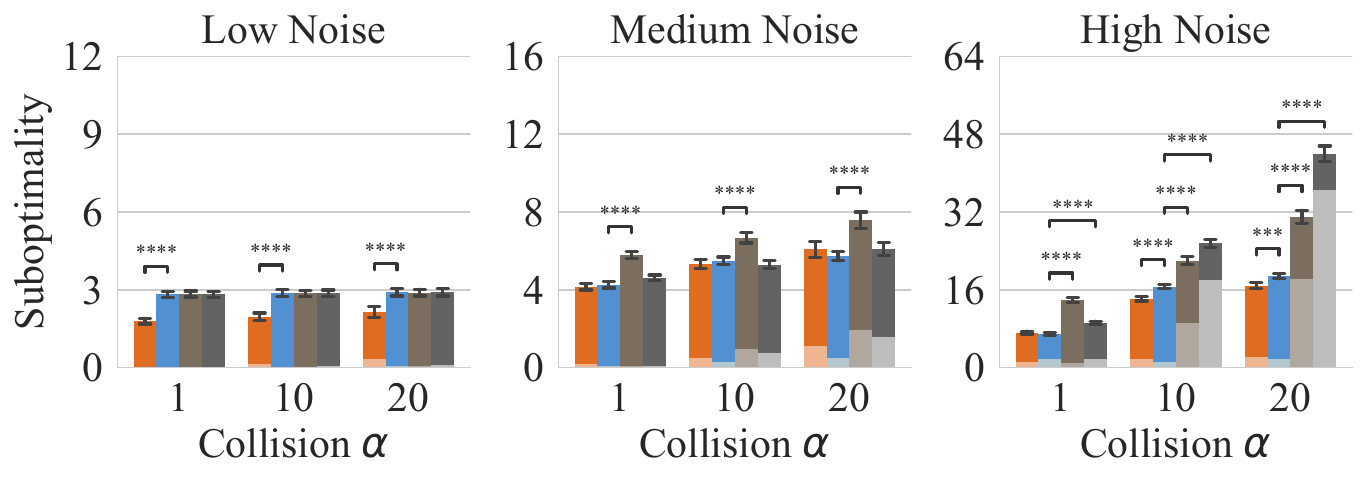}
            \caption{Forest}
            \label{fig:bar_dirtfish}
        \end{subfigure}
        \begin{subfigure}[t]{\linewidth}
            \centering
            \includegraphics[width=\linewidth]{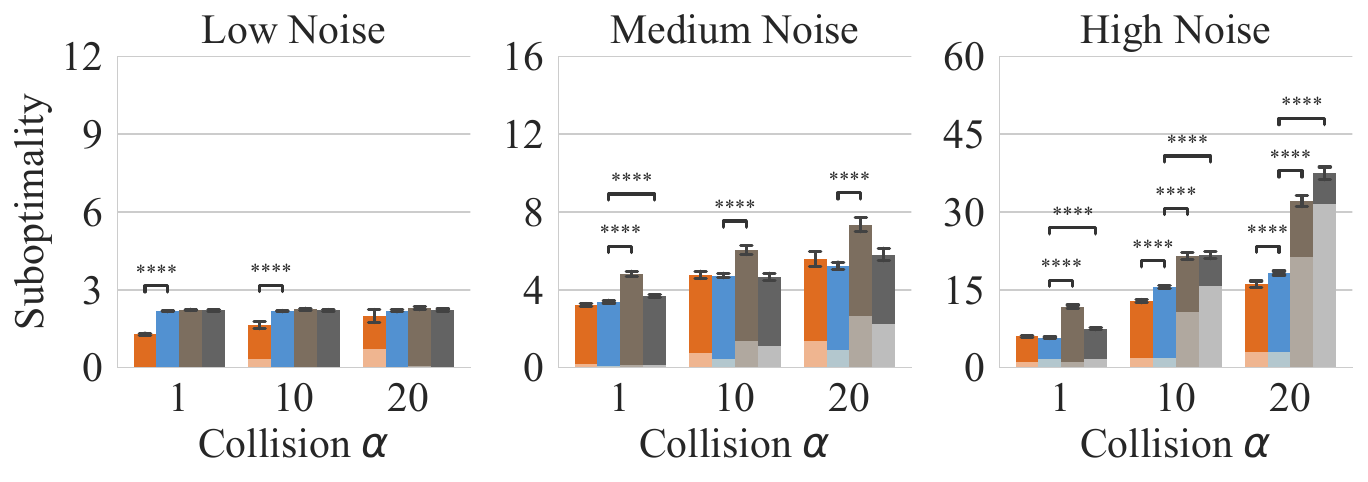}
            \caption{Desert}
            \label{fig:bar_ellensburg}
        \end{subfigure}
     \end{subfigure}
    \hfill
     \begin{subfigure}[t]{0.49\linewidth}
        \begin{subfigure}[t]{\linewidth}
            \centering
            \includegraphics[width=\linewidth]{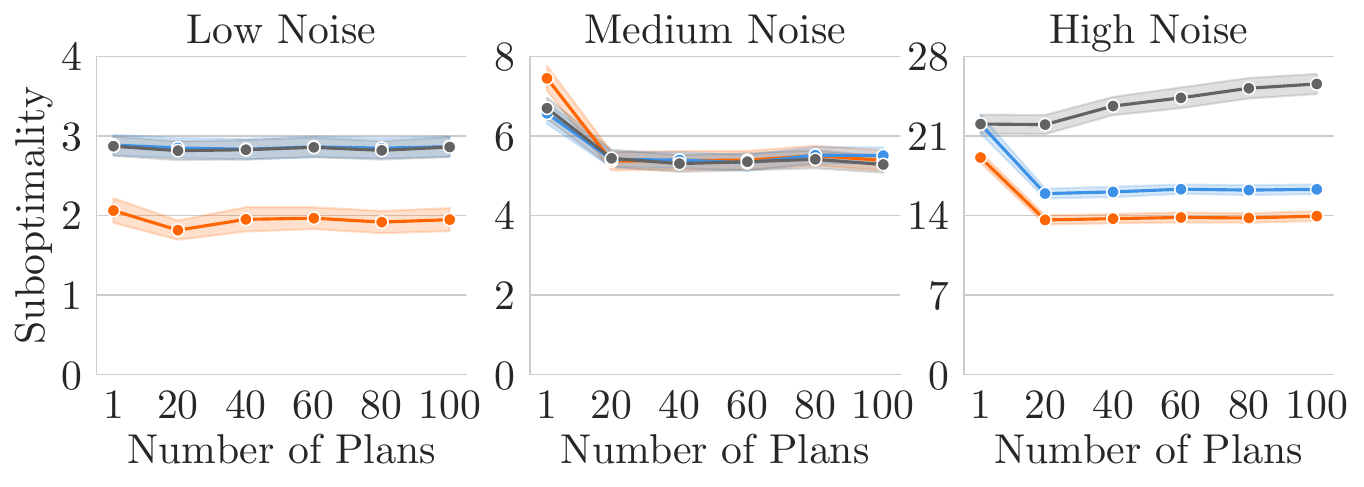}
            \caption{Paths}
            \label{fig:ablation_paths}
        \end{subfigure}
        \begin{subfigure}[t]{\linewidth}
            \centering
            \includegraphics[width=\linewidth]{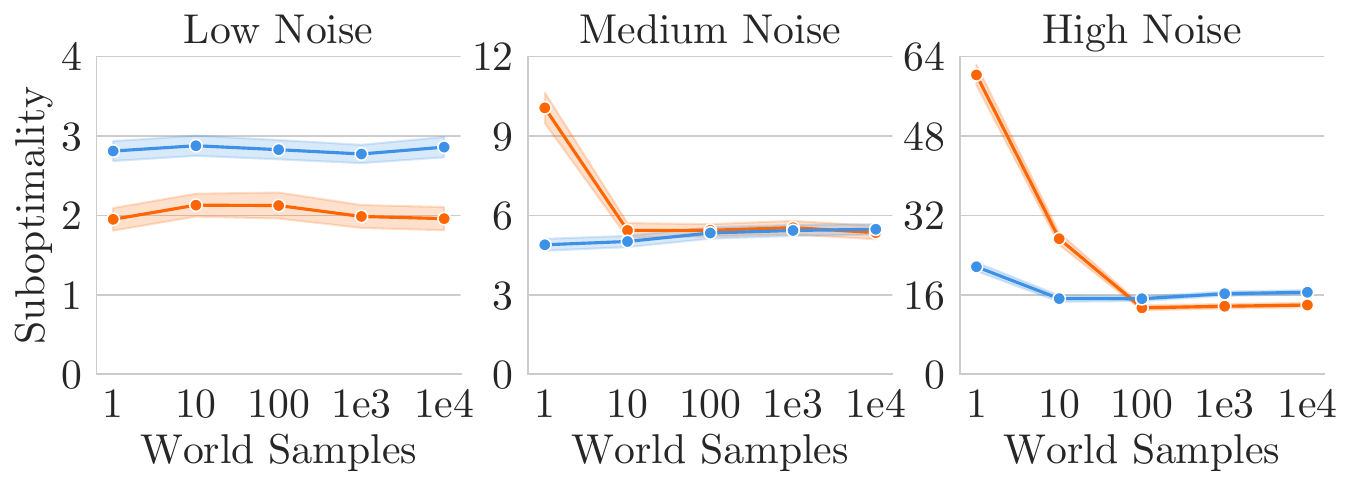}
            \caption{World Samples}
            \label{fig:ablations_samples}
        \end{subfigure}
     \end{subfigure}
    \caption{
    \emph{Left:} Suboptimality plots for (a) Forest and (b) Desert datasets.
    We perform a Welch's t-test for difference of means, with a Bonferroni correction of 90 for all pairwise comparisons involving DREAMS.
    $\text{*}: p < 0.01, \text{**}: p < 0.001, \text{***}: p < 0.0001, \text{****}: p < 0.00001$.
    \emph{Right:} Ablation study for varying (c) number of sampled plans and (d) number of world samples in evaluation.
    (Error bars in both figures denote 95\% confidence intervals.)
    }
    \label{fig:main_plot}
    \vspace{-1\baselineskip}
\end{figure*}

\begin{figure}
    \centering
    \includegraphics[width=\linewidth]{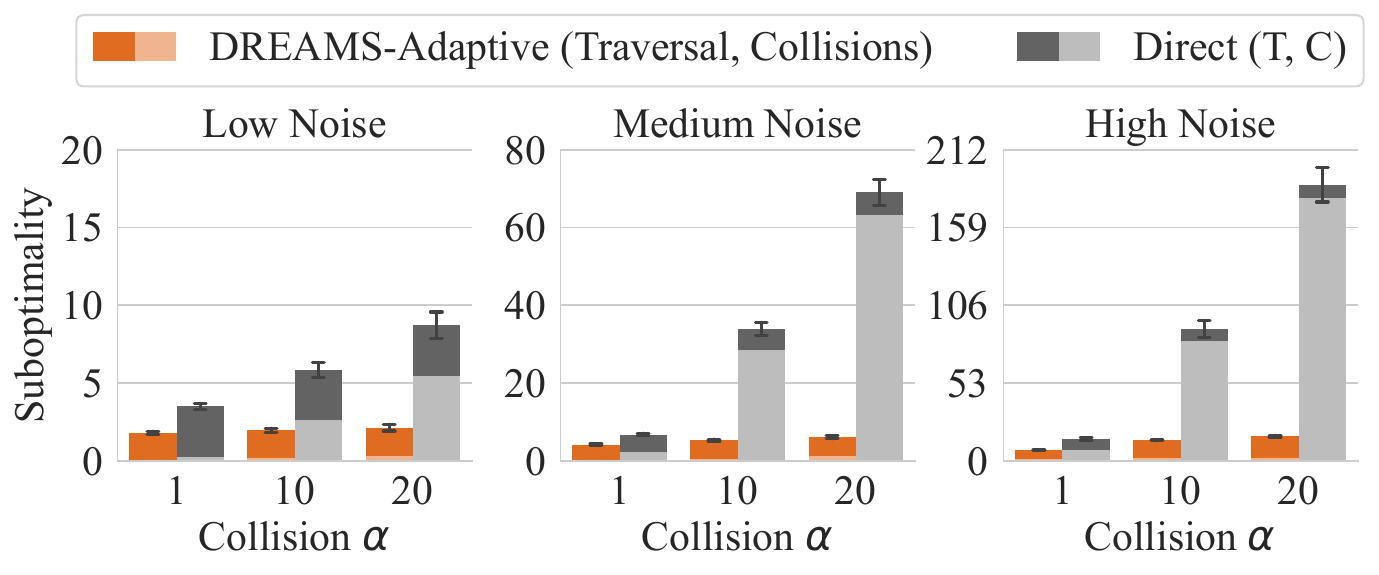}
    \caption{Comparison of DREAMS to Direct on the more challenging Forest dataset. Direct incurs many more collisions because it does not reason about the likelihood of the plans directly. Note: The suboptimality axis is much higher than Fig. \ref{fig:bar_dirtfish}.}
    \label{fig:direct_barplot}
\end{figure}


\subsection{Results}
Fig. \ref{fig:bar_dirtfish} and \ref{fig:bar_ellensburg} show suboptimality results for the Forest and Desert datasets.
An ablation study with the more challenging Forest dataset is visualized in  Fig. \ref{fig:ablation_paths} and \ref{fig:ablations_samples}.

Qualitative results of traversed paths on the final posterior are shown in Fig. \ref{fig:qualitative}.
Fig. \ref{fig:comparison} compares each algorithm under the exact same scenarios. Table \ref{table:plan_time} gives planning time results.

\textbf{H1.} In both datasets (Fig. \ref{fig:bar_dirtfish} and \ref{fig:bar_ellensburg}), DREAMS-Fixed is competitive with DRPS and Sampled A* in Low noise. It is either competitive or outperforms them in Medium noise. In High noise, it dominates for all $\alpha$ values tested. H1 is supported.
It is not surprising that all algorithms achieve similar results in Low noise, as the sampled plans will likely be near-optimal (i.e., most sampled worlds are close to the true world).
In Medium noise, we observe the benefit of multiple samples as both DREAMS variants and Sampled A* perform better than DRPS. But in High noise, Sampled A* incurs a high collision cost as it does not explicitly reason about collisions; all plans seem equally likely because the distribution of plans is spread out, showing less benefit to multiple samples without explicit collision reasoning. Fig. \ref{fig:comparison}, bottom shows an example scenario of this behavior.

\textbf{H2.} In Fig. \ref{fig:bar_dirtfish} and \ref{fig:bar_ellensburg}, we see a mixed result between DREAMS variants: DREAMS-Adaptive has statistically-significant lower cost in some cases but not all. With Low noise, DREAMS-Adaptive can move faster reducing its traversal time without incurring too much collision cost. In Medium noise, DREAMS-Adaptive and DREAMS-Fixed achieve similar performance; this suggests that these higher speeds do not properly balance speed and safety. In High noise, DREAMS-Adaptive seems to trade-off collisions and traversal better. Because the results are mixed, H2 is not strongly supported.

\begin{table}[h!]
    \vspace{5pt}
    \centering
    \setlength{\tabcolsep}{4pt} 
    \begin{tabular}{lccc}
    \toprule
        Algorithm       & Proposer Time & Acceptor Time  & Total Time    \\ \midrule
        DREAMS-Adaptive & $1.55\pm0.09$ & $1.59\pm0.01$  & $3.14\pm0.09$ \\
        DREAMS-Fixed    & $1.54\pm0.09$ & $0.33\pm0.00$  & $1.88\pm0.09$ \\
        DRPS            & $0.02\pm0.00$ & $0.01\pm0.00$  & $0.03\pm0.00$ \\
        Sampled A*      & $1.47\pm0.08$ & $1.12\pm0.02$  & $2.58\pm0.08$ \\
        \bottomrule
    \end{tabular}
    \caption{Planning time evaluations for each algorithm, on the same set of 300 evaluation runs.
    \emph{Proposer:} DREAMS and Sampled A* plan 100 paths per iteration, while DRPS plans a single path. Because there is no dependency between planning each independent posterior sample, this can be easily accelerated by parallelization (results sequential).
    \emph{Acceptor:} Since DRPS only proposes one path, the acceptor time is negligible. DREAMS-A evaluates five speeds taking more time while DREAMS-F evaluates just one. Aggregating edge centrality across plans is also relatively expensive for Sampled A*.
    }
\label{table:plan_time}
\vspace{-1\baselineskip}
\end{table}

\textbf{H3.} Fig. \ref{fig:ablation_paths} shows that more sampled plans reduces the total cost across multiple noise values for DREAMS. The leveling off at 20 plans shows that the sampled set is fairly representative of our posterior distribution. H3 is supported. Surprisingly, Sampled A* performed worse with increased samples at High noise. We attribute this to High noise causing a spread out distribution of plans where no plan has a large mean centrality, resulting in near random choice (See Fig. \ref{fig:comparison}, bottom). Increasing the number of plans introduces more options for Sampled A* to choose from; if these are likely to be in collision, this will generally increase the cost incurred.

\textbf{H4.} Fig. \ref{fig:ablations_samples} shows more world samples in evaluation improves performance for DREAMS-Adaptive at Medium and High noise. For DREAMS-Fixed, we only see change in High noise. In this setting, it suggests a few samples captures the true cost well in Low and Medium noise, but more samples are needed in High noise. H4 is supported for High noise.

\textbf{H5.} Fig. \ref{fig:direct_barplot} shows the suboptimality comparison between DREAMS-Adaptive and Direct. Direct incurs a very high collision cost and total suboptimality because it finds paths with a high likelihood of collisions. We attributed this to the probability of collision being a multiplicative factor instead of an actual probability, meaning Direct may choose plans that are highly unlikely but have a low total cost. H5 is supported.

\section{Discussion}
\label{sec:discussion}
There remain multiple limitations and avenues for future work. DREAMS currently relies on a hand-tuned cost function for its evaluation step. While we show it works in our setting with traversal time and collisions, if an application considers more costs it could quickly become difficult to design a good cost function. A learned cost function using ground truth information or demonstrations may be a more scalable alternative. Second, we perform sampling on a graph posterior that is obtained through taking the maximum probability of collision along an edge. While this heuristic can give a nice distribution of paths, it is worth exploring other sampling methods like directly sampling costmaps from the perception model or sampling paths from a neural planner.

\section{Disclaimer}
The views, opinions and/or findings
expressed are those of the author and should not be interpreted as
representing the official views or policies of the Department of Defense or
the U.S. Government.

{
\small
\bibliographystyle{IEEEtran}
\bibliography{ref}

\begin{thebibliography}{10}
\providecommand{\url}[1]{#1}
\csname url@rmstyle\endcsname
\providecommand{\newblock}{\relax}
\providecommand{\bibinfo}[2]{#2}
\providecommand\BIBentrySTDinterwordspacing{\spaceskip=0pt\relax}
\providecommand\BIBentryALTinterwordstretchfactor{4}
\providecommand\BIBentryALTinterwordspacing{\spaceskip=\fontdimen2\font plus
\BIBentryALTinterwordstretchfactor\fontdimen3\font minus
  \fontdimen4\font\relax}
\providecommand\BIBforeignlanguage[2]{{%
\expandafter\ifx\csname l@#1\endcsname\relax
\typeout{** WARNING: IEEEtran.bst: No hyphenation pattern has been}%
\typeout{** loaded for the language `#1'. Using the pattern for}%
\typeout{** the default language instead.}%
\else
\language=\csname l@#1\endcsname
\fi
#2}}

\bibitem{racer}
``Robotic autonomy in complex environments with resiliency (racer),''
  \url{https://www.darpa.mil/program/robotic-autonomy-in-complex-environments-with-resiliency}.

\bibitem{papadimitriou1987}
C.~H. Papadimitriou and J.~N. Tsitsiklis, ``The complexity of markov decision
  processes,'' \emph{Math. Oper. Res.}, 1987.

\bibitem{hou2023}
B.~Hou and S.~Srinivasa, ``Dynamic replanning with posterior sampling,'' in
  \emph{{IEEE/RSJ} International Conference on Intelligent Robots and Systems},
  2023.

\bibitem{somani2013}
A.~Somani, N.~Ye, D.~Hsu, and W.~S. Lee, ``Despot: Online pomdp planning with
  regularization,'' in \emph{Advances in Neural Information Processing
  Systems}, 2013.

\bibitem{lim2017}
Z.~W. Lim, D.~Hsu, and W.~S. Lee, ``Shortest path under uncertainty :
  Exploration versus exploitation,'' in \emph{Conference on Uncertainty in
  Artificial Intelligence}, 2017.

\bibitem{yoon2007}
S.~Yoon, A.~Fern, and R.~Givan, ``Ff-replan: A baseline for probabilistic
  planning,'' in \emph{International Conference on Automated Planning and
  Scheduling}, 2007.

\bibitem{stentz1994}
A.~Stentz, ``The d* algorithm for real-time planning of optimal traverses,''
  Carnegie Mellon University, Pittsburgh, PA, Tech. Rep., 1994.

\bibitem{koenig2002}
S.~Koenig and M.~Likhachev, ``D*lite,'' in \emph{{AAAI} Conference on
  Artificial Intelligence}, 2002.

\bibitem{ferguson2005}
D.~Ferguson and A.~Stentz, ``Field d*: An interpolation-based path planner and
  replanner,'' in \emph{International Symposium on Robotics Research}, 2005.

\bibitem{ross2008}
S.~Ross, J.~Pineau, S.~Paquet, and B.~Chaib-draa, ``Online planning algorithms
  for pomdps,'' \emph{J. Artif. Int. Res.}, 2008.

\bibitem{silver2010}
D.~Silver and J.~Veness, ``Monte-carlo planning in large pomdps,'' in
  \emph{Advances in Neural Information Processing Systems}, 2010.

\bibitem{sunberg2017}
Z.~Sunberg and M.~J. Kochenderfer, ``Online algorithms for pomdps with
  continuous state, action, and observation spaces,'' in \emph{International
  Conference on Automated Planning and Scheduling}, 2017.

\bibitem{chen2016}
M.~Chen, E.~Frazzoli, and D.~Hsu, ``Pomdp-lite for robust robot planning under
  uncertainty,'' in \emph{{IEEE} International Conference on Robotics and
  Automation}, 2016.

\bibitem{papadimitriou1989}
C.~H. Papadimitriou and M.~Yannakakis, ``Shortest paths without a map,'' in
  \emph{Automata, Languages and Programming}, 1989.

\bibitem{eyerich2010}
P.~Eyerich, T.~Keller, and M.~Helmert, ``High-quality policies for the canadian
  traveler’s problem,'' \emph{{AAAI} Conference on Artificial Intelligence},
  2010.

\bibitem{dey2014}
D.~Dey, A.~Kolobov, R.~Caruana, E.~Kamar, E.~Horvitz, and A.~Kapoor, ``Gauss
  meets canadian traveler: Shortest-path problems with correlated natural
  dynamics,'' in \emph{International Conference on Autonomous Agents and
  Multi-agent Systems}, 2014.

\bibitem{yoon2008}
S.~Yoon, A.~Fern, R.~Givan, and S.~Kambhampati, ``Probabilistic planning via
  determinization in hindsight,'' in \emph{{AAAI} Conference on Artificial
  Intelligence}, 2008.

\bibitem{chung2019}
J.~J. Chung, A.~J. Smith, R.~Skeele, and G.~A. Hollinger, ``Risk-aware graph
  search with dynamic edge cost discovery,'' \emph{International Journal of
  Robotics Research}, 2019.

\bibitem{barbosa2021}
F.~S. Barbosa, B.~Lacerda, P.~Duckworth, J.~Tumova, and N.~Hawes, ``Risk-aware
  motion planning in partially known environments,'' \emph{Conference on
  Decision and Control}, 2021.

\bibitem{suresh2019}
A.~Suresh and S.~Mart{\'{\i}}nez, ``Planning under risk and uncertainty based
  on prospect-theoretic models,'' \emph{arXiv preprint arXiv:1904.02851}, 2019.

\bibitem{feyzabadi2014}
S.~Feyzabadi and S.~Carpin, ``Risk-aware path planning using hirerachical
  constrained markov decision processes,'' in \emph{{IEEE} International
  Conference on Automation Science and Engineering}, 2014.

\bibitem{murphy2013}
L.~Murphy and P.~Newman, ``Risky planning on probabilistic costmaps for path
  planning in outdoor environments,'' \emph{{IEEE} Transactions on Robotics},
  2013.

\bibitem{cai2022}
X.~Cai, M.~Everett, J.~Fink, and J.~P. How, ``Risk-aware off-road navigation
  via a learned speed distribution map,'' in \emph{{IEEE/RSJ} International
  Conference on Intelligent Robots and Systems}, 2022.

\bibitem{uryasev2001}
S.~Uryasev and R.~T. Rockafellar, \emph{Conditional Value-at-Risk: Optimization
  Approach}.\hskip 1em plus 0.5em minus 0.4em\relax Springer US, 2001.

\bibitem{asmuth2012}
J.~Asmuth, L.~Li, M.~L. Littman, A.~Nouri, and D.~Wingate, ``A bayesian
  sampling approach to exploration in reinforcement learning,''
  \emph{Conference on Uncertainty in Artificial Intelligence}, 2012.

\bibitem{strens2000}
M.~J.~A. Strens, ``A bayesian framework for reinforcement learning,'' in
  \emph{International Conference on Machine Learning}, 2000.

\bibitem{wilson2007}
A.~Wilson, A.~Fern, S.~Ray, and P.~Tadepalli, ``Multi-task reinforcement
  learning: A hierarchical bayesian approach,'' in \emph{International
  Conference on Machine Learning}, 2007.

\bibitem{thompson1933}
W.~R. Thompson, ``On the likelihood that one unknown probability exceeds
  another in view of the evidence of two samples,'' \emph{Biometrika}, 1933.

\bibitem{freeman1977}
L.~C. Freeman, ``A set of measures of centrality based on betweenness,''
  \emph{Sociometry}, 1977.

\bibitem{shaban2022}
A.~Shaban, X.~Meng, J.~Lee, B.~Boots, and D.~Fox, ``Semantic terrain
  classification for off-road autonomous driving,'' in \emph{{IEEE}
  International Conference on Robotics and Automation}, 2022.

\bibitem{schmid2022}
R.~Schmid, D.~Atha, F.~Schöller, S.~Dey, S.~Fakoorian, K.~Otsu, B.~Ridge,
  M.~Bjelonic, L.~Wellhausen, M.~Hutter, and A.~Agha-mohammadi,
  ``Self-supervised traversability prediction by learning to reconstruct safe
  terrain,'' in \emph{{IEEE/RSJ} International Conference on Intelligent Robots
  and Systems}, 2022.

\end{thebibliography}
}



\end{document}